\pdfoutput=1

\documentclass[11pt]{article}

\usepackage[]{acl}

\usepackage{times}
\usepackage{latexsym}

\usepackage[T1]{fontenc}

\usepackage[utf8]{inputenc}

\usepackage{microtype}

\usepackage{paralist}
\usepackage{style/style}
\usepackage{amssymb}
\usepackage{pifont}
\newcommand{\cmark}{\ding{51}}%
\newcommand{\xmark}{\ding{55}}%

\usepackage{booktabs}
\usepackage{multirow}
\usepackage{rotating}
\usepackage{tipa}

\usepackage[normalem]{ulem}
\usepackage{comment}

\title{An Analysis of Negation in Natural Language Understanding Corpora}

\author{Md Mosharaf Hossain\mbox{\normalfont,}\textsuperscript{\textipa{8}}
        Dhivya Chinnappa\mbox{\normalfont,}\textsuperscript{\textipa{U}} \and        
        Eduardo Blanco\textsuperscript{\textipa{7}}\\
\textsuperscript{\textipa{8}}Department of Computer Science and Engineering, University of North Texas\\
\textsuperscript{\textipa{U}}Thomson Reuters\\
\textsuperscript{\textipa{7}}School of Computing and Augmented Intelligence, 	Arizona State University\\
{\footnotesize
\texttt{mdmosharafhossain@my.unt.edu} \hspace{.2cm}
\texttt{dhivya.infant@gmail.com} \hspace{.2cm}
\texttt{eduardo.blanco@asu.edu}}
}

\begin{document}
\maketitle
\begin{abstract}
This paper analyzes negation in eight popular corpora spanning six natural language understanding tasks.
We show that these corpora have few negations compared to general-purpose English, 
and that the few negations in them are often unimportant.
Indeed, one can often ignore negations and still make the right predictions.
Additionally, experimental results show that state-of-the-art transformers trained with these corpora obtain substantially worse results with instances that contain negation, especially if the negations are important.
We conclude that new corpora accounting for negation are needed to solve natural language understanding tasks when negation is present.
\end{abstract}

\section{Introduction}
\label{s:introduction}
Natural language understanding (NLU) is an umbrella term used to refer to 
any task that requires text understanding. 
For example,
question answering~\cite{rajpurkar-etal-2016-squad},
information extraction~\cite{stanovsky-etal-2018-supervised},
coreference resolution~\cite{wu-etal-2020-corefqa}, and
machine reading~\cite{yang-etal-2019-enhancing-pre}, among many others,
are tasks that fall under natural language understanding.
The threshold for claiming that a system understands natural language is ever-moving.
New corpora are often justified by pointing out that state-of-the-art models do not obtain good results.
After years of  steady improvements, more powerful models eventually obtain so-called human performance,
and at that point new, more challenging corpora are created.

Many corpora for natural language understanding tasks
contain language generated by annotators rather than retrieved from texts written independently of the corpus creation process.
These corpora are certainly useful and have facilitated tremendous progress.
Annotator-generated examples, however,
carry the risk of evaluating systems with synthetic language that is not representative of language in the wild.
For example, annotators are likely to use negation when asked to write a text that contradicts something
despite contradictions in the wild need not have a negation~\cite{gururangan-etal-2018-annotation}.
Recently,
\newcite{kwiatkowski-etal-2019-natural} present a large corpus for question answering
that consists of natural questions (\ie{}, asked by somebody with a real information need)
in order to encourage research in a more realistic scenario.
This contrasts with previous corpora, where the questions were written by annotators after being told the answer~\cite{rajpurkar-etal-2016-squad}.


In this paper, we explore the role of negation in eight corpora for six popular natural language understanding tasks.
Our goal is to check whether negation plays the role it deserves in these tasks.
To our surprise, we conclude that negation is virtually ignored by answering the following questions:\footnote{Code and data available at \url{https://github.com/mosharafhossain/negation-and-nlu}.}

\begin{compactenum}
    \item Do NLU corpora contain as many negations as general-purpose texts?~(they don't);
    \item Do the (few) negations in NLU corpora play a role in solving the tasks?~(they don't); and
    \item Do state-of-the-art transformers trained with NLU corpora face challenges with instances that contain negation?
    (they do, especially if the negation is important).
\end{compactenum}

\section{Background and Related Work}
\label{s:background}
We work with the eight corpora covering six tasks summarized below and exemplified in Table \ref{t:nlu_examples}.

We select two corpora for question answering:
CommonsenseQA~\cite{talmor-etal-2019-commonsenseqa} and COPA~\cite{roemmele2011choice}.
CommonsenseQA consists of multi-choice questions (5 candidate answers) that require some degree of commonsense.
COPA presents a premise (\eg{}, \emph{The man broke his toe}) and a question (\eg{}, \emph{What was the cause of this?})
and the system must choose between two plausible alternatives
(\eg{} \emph{He got a hole in his sock} or \emph{He dropped a hammer on his foot}).

For textual similarity and paraphrasing, we select
QQP\footnote{https://www.quora.com/q/quoradata/First-Quora-Dataset-Release-Question-Pairs} and
STS-B~\cite{cer-etal-2017-semeval}.
QQP consists of pairs of questions and the task is to determine whether they are paraphrases.
STS-B consists of pairs of texts and the task is to determine how semantically similar they are with a score from~0~to~5.

We select one corpus for the remaining tasks.
For inference, we work with QNLI~\cite{rajpurkar-etal-2016-squad},
which consists in determining whether a text is a valid answer to a question.
We use WiC~\cite{pilehvar-camacho-collados-2019-wic} for word sense disambiguation.
WiC consists in determining whether two instances of the same word~(in
two sentences; italicized in Table~\ref{t:nlu_examples}) are used with the same meaning.
For coreference resolution, we choose
WSC~\cite{10.5555/3031843.3031909}, which consists in determining whether a pronoun and a noun phrase are co-referential (italicized in Table~\ref{t:nlu_examples}).
Finally, we work with SST-2~\cite{socher-etal-2013-recursive} for sentiment analysis.
The task consists in determining whether a sentence from a collection of movie reviews has positive or negative sentiment.

For convenience, we work with the formatted versions of these corpora in the
GLUE \cite{wang-etal-2018-glue} and SuperGLUE \cite{NIPS2019_8589} benchmarks.
The only exception is CommonsenseQA, which is not part of these benchmarks.

\noindent
\textbf{Related Work}
\label{ss:related_work}
Previous work has shown that
SNLI~\cite{bowman-etal-2015-large}
and MNLI~\cite{N18-1101}
have annotation artifacts~(\eg{}, negation is a strong indicator of \emph{contradictions})~\cite{gururangan-etal-2018-annotation}.
The literature has also shown that simple adversarial attacks including negation cues are very
effective~\cite{naik-etal-2018-stress,wallace-etal-2019-universal}. 
\newcite{kovatchev-etal-2019-qualitative} analyze 11 paraphrasing systems and show that they obtain substantially worse results when negation is present. 

More recently, \newcite{ribeiro-etal-2020-beyond}
show that negation is one of the linguistic phenomena commercial sentiment analysis struggle with.
Several previous works have investigated the (lack of) ability of transformers to make inferences when negation is present.
For example,
\newcite{ettinger-2020-bert} conclude that BERT is unable to complete sentences when negation is present.
BERT also faces challenges
solving the task of natural language inference (\ie{},
identifying entailments and contradictions)
with monotonicity and negation ~\cite{
geiger-etal-2020-neural,
yanaka-etal-2019-help}.
\newcite{warstadt-etal-2019-investigating} show the limitations of BERT making acceptability judgments with sentences that contain negative polarity items.
Most related to out work,
\newcite{hossain-etal-2020-analysis} analyze the role of negation in three natural language inference corpora:
RTE~\cite{dagan:2005:PRT:2100045.2100054,
bar2006second,
giampiccolo-etal-2007-third,
bentivogli2009fifth},
SNLI and MNLI.
In this paper, we present a similar analysis,
but we move beyond natural language inference and
work with eight corpora spanning six natural language understanding tasks.

\begin{table}
\small
\centering
\begin{tabular}{lrr}
\toprule
 & \#sents. & \% w/ neg. \\
\midrule

Question Answering \\
~~~CommonsenseQA & 12,102  & 14.5 \\
~~~COPA          & 1,000   &  0.8 \\ 
\midrule

Similarity and Paraphrasing \\
~~~QQP   & 1,590,482  & 8.1 \\
~~~STS-B & 17,256     & 7.1 \\ 
\midrule

Inference \\
~~~ QNLI & 231,338    & 8.7 \\
\midrule

Word Sense Disambiguation \\
~~~ WiC & 14,932    & 8.2 \\
\midrule

Coreference Resolution \\
~~~ WSC & 804       & 26.2 \\
\midrule

Sentiment Analysis \\
~~~ SST-2 & 70,042    & 16.0 \\

\midrule \midrule

General-purpose English \\
~~~ all sentences &
8,300,000 & 22.6--29.9 \\
~~~ only questions &
456,214 & 15.8--20.2
 
 \\ \bottomrule

\end{tabular}

\caption{Number of sentences and percentage of sentences containing negation
  in natural language understanding corpora. 
  All but WSC contain substantially fewer negations than general-purpose English texts.
  }
\label{t:neg_counts_english}
\end{table}

\begin{table*}[ht!]
\small
\centering
\begin{tabular}{cp{4.5in}cc}
\toprule
& Example & & Important? \\ \midrule
\multirow{4}{*}{\begin{sideways}CmmsnsQA\end{sideways}}
& [\ldots] he (John) \uline{never} saw the lady before. They were what?
  & \multirow{2}{*}{C} & \multirow{2}{*}{\cmark} \\ 
& A) pay debts, B) slender, C) unacquainted, D) free flowing, E) sparse \\ \cmidrule{2-4} 

& When you travel you should what in case of  \uline{unexpected} costs?
  & \multirow{2}{*}{E} & \multirow{2}{*}{\xmark} \\ 
& A) go somewhere,  B) energy,  C) spend frivilously,  D) fly in airplane, E) have money \\ \midrule

\multirow{4}{*}{\begin{sideways}QQP\end{sideways}}           
& What are some \uline{not}-so-boring baby shower games ?                                                                                                                                                                                                               & \multirow{2}{*}{yes}              & \multirow{2}{*}{\cmark}   \\ 
                               & What are some baby shower games that are actually fun?                                                                                                                                                                                                                                      &                                 &                        \\ \cmidrule{2-4} 
                               & Who was philosophical guru of Shivaji Maharaj?                                                                                                                                                                                                                                              & \multirow{2}{*}{no}              & \multirow{2}{*}{\xmark} \\ 
                               & What are the \uline{unknown} facts of shivaji maharaj?                                                                                                                                                                                                                                              &                                 &                        \\ \midrule

\multirow{4}{*}{\begin{sideways}STS-B\end{sideways}}         
                               & Colin Powell, the Secretary of State, said contacts with Iran would \uline{not} stop. & \multirow{2}{*}{4.3}            & \multirow{2}{*}{\cmark} \\ 
                               & Secretary of State Colin Powell said yesterday that contacts with Iran would continue. &                                 &                        \\ \cmidrule{2-4}

& Well for one a being could have a \uline{non}-physical existance and yet \uline{not} even be in your mind.
& \multirow{2}{*}{3.4}            & \multirow{2}{*}{\xmark}   \\ 
                               &
                               The difference is huge, as \uline{not} all \uline{non}-physical things exist in minds.                              
&                                 &                        \\ \midrule

\multirow{6}{*}{\begin{sideways}QNLI\end{sideways}}          & Who did BSkyB team up with as it was \uline{not} part of consortium?                                                                                                                                                                                                                       & \multirow{2}{*}{yes}     & \multirow{2}{*}{\cmark}   \\ 
                               & While BSkyB had been excluded from being a part of the [\ldots], BSkyB was able to join ITV Digital's free-to-air replacement, Freeview, in which it holds an equal stake [\ldots]  &                                 &                        \\ \cmidrule{2-4} 
                               & In what year did Lavoisier publish his work on combustion?                                                                                                                                                                                                                                  & \multirow{2}{*}{no} & \multirow{2}{*}{\xmark} \\ 
                               & In one experiment, Lavoisier observed that there was \uline{no} overall increase in weight when tin and air were heated in a closed container. &                                 &                        \\ \midrule

\multirow{2}{*}{\begin{sideways}SST-2\end{sideways}}          & It's \uline{not} the ultimate depression-era gangster movie.                                                                                                                                                                                                                                      & neg.                               & \cmark                    \\ \cmidrule{2-4}
                               & Whaley's determination to immerse you in sheer, \uline{unrelenting} wretchedness is exhausting.                                                                                                                                                                                                  & neg.                               & \xmark                  \\ \midrule \midrule

\multirow{2}{*}{\begin{sideways}WiC\end{sideways}}           & The \emph{intention} of this legislation is to boost the economy.                                                                                                                                                                                                                                  & \multirow{2}{*}{same}              & \multirow{2}{*}{\xmark} \\ 
                               & Good \emph{intentions} are \uline{not} enough.                                                                                                                                                                                                                                                             &                                 &                        \\ \midrule

\multirow{2}{*}{\begin{sideways}WSC\end{sideways}} & \emph{Sam and Amy} are passionately in love, but Amy's parents are \uline{unhappy} about it, because \emph{they}~are only fifteen.                                                                                                                                                                                                                                             & \multirow{2}{*}{yes}                               & \multirow{2}{*}{\xmark}                  \\ \bottomrule

\end{tabular}

\caption{Examples containing negation (underlined) from the validation datasets of the natural language understanding corpora we work with. 
The third column presents the expected answer for the example (a choice, judgment, or score depending on the task).
The last column indicates whether the negation is important.
}
\label{t:nlu_examples}
\end{table*}

\section{Research Questions and Analysis}
\label{s:experiments}

\noindent
\textbf{Q1: Do natural language understanding corpora contain as many negations as general-purpose English texts?}
In order to automatically identify negation cues,
we train a negation cue detector with the largest corpus available, ConanDoyle-neg~\cite{morante-daelemans-2012-conandoyle}.
The cue detector is based on the RoBERTa pretrained language model~\cite{liu2019roberta};
we provide details about the architecture and training process in Appendix \ref{s:negation_detection}.
Our cue detector obtains the best results to date:
F1:~93.79 vs. 92.94 \cite{khandelwal-sawant-2020-negbert}.
ConanDoyle-neg (and thus our cue detector) identifies
common negation cues such as \emph{no}, \emph{not}, \emph{n't} and \emph{never},
affixal negation cues such as \emph{impossible} and \emph{careless},
and lexical negations such as \emph{deny} and \emph{avoid}.

Table \ref{t:neg_counts_english} presents the percentage of sentences that contain negation in
(a) the eight corpora we work with 
and
(b) general-purpose English. 
We take the latter percentage (all sentences) from \newcite{hossain-etal-2020-analysis},
who run a negation cue detector in online reviews, conversations, and books.
Additionally, we also present the percentages in questions.
Negation is much less common in 
all natural language understanding corpora but WSC (0.8\%--16\%)
than in general-purpose English~(22.6\%--29.9\%).
Note that negation is also underrepresented in corpora that primarily contain questions (general-purpose: 15.8\%--20.2\%; COPA: 0.8\%, QQP: 8.1\%).

\begin{table*}[ht!]
\small
\centering
\begin{tabular}{ccp{4.5in}cc}
\toprule
& & Example & & Important? \\ \midrule
\multirow{9}{*}{\begin{sideways}CommonsenseQA\end{sideways}}
& \multirow{5}{*}{\begin{sideways}Syntactic\end{sideways}}
& Where would a person live if they wanted \uline{no} neighbors?
  & \multirow{1}{*}{D} & \multirow{1}{*}{\cmark} \\ 
& & A) housing estate, B) neighborhood, C) mars, D) woods, E) suburbs \\ \cmidrule{3-5}  

& & The teacher does\uline{n't} tolerate noise during a test in their what?
  & \multirow{1}{*}{E} & \multirow{1}{*}{\xmark} \\ 
& & A) movie theatre, B) bowling alley, C) factory, D) store, E) classroom \\ \cmidrule{2-5} 

& \multirow{5}{*}{\begin{sideways}Morpho.\end{sideways}}
& What might result in an \uline{unsuccessful} suicide attempt?
  & \multirow{1}{*}{B} & \multirow{1}{*}{\cmark} \\ 
& & A) die,  B) interruption,  C) bleed,  D) hatred, E) dying \\  \cmidrule{3-5} 

& & How are the conditions for someone who is living in a \uline{homeless} shelter?
  & \multirow{1}{*}{A} & \multirow{1}{*}{\xmark} \\ 
& & A) sometimes bad,  B) happy,  C) respiration,  D) growing older, E) death \\
\midrule

\multirow{9}{*}{\begin{sideways}STS-2\end{sideways}}
& \multirow{5}{*}{\begin{sideways}Syntactic\end{sideways}}
& Despite the evocative aesthetics evincing the hollow state of modern love life, the film \uline{never} percolates beyond a monotonous whine.
  & neg. & \multirow{1}{*}{\cmark} \\ \cmidrule{3-5} 

& & Even if you do\uline{n't} think (kissinger's) any more guilty of criminal activity than most contemporary statesmen, he'd sure make a courtroom trial great fun to watch.
  & pos. & \multirow{1}{*}{\xmark} \\  \cmidrule{2-5} 

& \multirow{3}{*}{\begin{sideways}Morpho.\end{sideways}}
& Makes for a pretty \uline{unpleasant} viewing experience.
  & neg. & \multirow{1}{*}{\cmark} \\ \cmidrule{3-5} 

& & For anyone \uline{unfamiliar} with pentacostal practices in general and theatrical phenomenon of hell houses in particular, it's an eye-opener .
  & pos. & \multirow{1}{*}{\xmark} \\ 
\bottomrule

\end{tabular}

\caption{Examples containing syntactic and morphological negation (underlined) from the validation datasets of CommonsenseQA and SST-2.
}
\label{t:nlu_examples_synt_morpho}
\end{table*}

\begin{table*}[ht!]
\small
\centering
\setlength{\tabcolsep}{.07in}
\begin{tabular}{lcccccccc}
\toprule
  & CmmnsnsQA & COPA  &  QQP  & STS-B & QNLI & WiC   &  WSC   & SST-2 \\
\midrule

validation w/o neg   & 0.60          & 0.73  &  0.90 & 0.92 / 0.91  & 0.93 & 0.67  &  0.63  & 0.94 \\
validation w/ neg    & 0.53          & n/a   &  0.91 & 0.85 / 0.84  & 0.91 & 0.64  &  0.59  & 0.93 \\
\midrule
~~~important (sample from Q2)   & 0.47      & n/a   &  0.73 & 0.57 / 0.62  & 0.67 & n/a   &  n/a   & 0.86 \\
~~~unimportant (sample from Q2) & 0.62      & n/a   &  0.92 & 0.85 / 0.84  & 0.92 & 0.64  &  0.59  & 0.95 \\

\bottomrule

\end{tabular}

\caption{
Results obtained with RoBERTa evaluating against
(a)~all instances with and without negation,
and
(b)~the sample of instances
 with negation we analyze in detail (important and unimportant).
 Since the datasets are unbalanced, we report macro F1-score for all tasks except STS-B, for which we report
 Pearson and Spearman correlations.
 Results are slightly lower with negation,
 and substantially lower with \emph{important} negations.
}
\label{t:results}
\end{table*}

\noindent
\textbf{Q2: Do the (few) negations in natural language understanding corpora play a role in solving the tasks?}
After showing that negation in underrepresented in natural language understanding corpora,
we explore whether the few negations they contain are important.
Given an instance from any of the corpora,
we consider a negation \emph{important}
if removing it changes the ground truth.
In other words, a negation is \emph{unimportant} if one can ignore it and still solve the task at hand.
Table \ref{t:nlu_examples} presents examples of important and unimportant negations.

We manually examine the negations in all instances containing negation from the validation split of each corpus
except QQP, for which we examine 1,000~(out of 5,196).
Note that COPA does not have any negations in the validation split, 
and many corpora have few 
instances containing negation
(CommonsenseQA: 184,
 STS-B: 225,
 QNLI: 852,
 WiC: 99,
 WSC: 52,
 and SST-2: 263).
We choose to work with the validation set because we want to compare results when negation is and is not important (Q3),
and the ground truth for the test splits of some corpora are not publicly available.

We observe that
(a)~all negations in WiC and WSC are unimportant,
and
(b)~the percentages of unimportant negations in CommonsenseQA, SST-2, QQP, STS-B, and QNLI are substantial:
45.1\%, 63\%, 97.4\%, 95.6\%, and 97.7\%, respectively. 
These percentages indicate that one can safely ignore (almost) all negations and still solve the benchmarks. 
Despite the fact that negations are not important in WSC and WiC, they do affect the experimental results (details in Q3).

We also analyze the role of two major types of negation:
syntactic (\emph{not}, \emph{no}, \emph{never}, etc.)
and morphological (i.e., affixes such as \emph{un-}, \emph{im-}, and \emph{-less}).
To this end, we work with CommonsenseQA and SST-2, which have lower percentages of unimportant negations (45.1\% and 63\%) than the other corpora we use (97.4\%--100\%). 
Table \ref{t:nlu_examples_synt_morpho} provides examples of these two negation types.
Perhaps unsurprisingly, syntactic negations are much more common than morphological negations~(CommonsenseQA: 88.6\% vs 11.4\%, SST-2: 71.9\% vs 28.1\%).
More importantly,
syntactic negations are more often important in SST-2 (42.3\% vs 23\%),
but both syntactic and morphological negation are roughly equaly important in CommonsenseQA (55.2\% vs 52.4\%). 

\noindent
\textbf{Q3: Do state-of-the-art transformers trained with NLU corpora face challenges with instances that contain negation?}
We conduct experiments with RoBERTa~\cite{liu2019roberta}.
More specifically, we use the implementation by~\newcite{phang2020jiant} and
train a model with the training split of each corpus. 
We refer the readers to the Appendix \ref{s:generated_neg_pairs} for the details about these models and hyperparameters.
We chose RoBERTa over other transformers because 4 out of the 10 best submissions to the SuperGLUE benchmark use it.\footnote{https://super.gluebenchmark.com/leaderboard}

Table \ref{t:results} presents the results evaluating the models with the corresponding validation splits.
RoBERTa obtains slightly worse results with the validation instances that have negation in all corpora;
the only exception is QQP (F1: 0.90 vs. 0.91).
These results lead to the  conclusion that negation \emph{may} only pose a small challenge to state-of-the-art transformers.

The results obtained evaluating with the important and unimportant negations from the samples analyzed in Question 2,
however, provide a different picture.
Indeed, we observe substantial drops in results in all tasks that have both kinds of negations.
More specifically,
we obtain 27\% lower results with instances containing important negations in QNLI~(F1:~0.92 vs.~0.67),
33\%/26\% lower in STS-B,
24\% lower in CommonsenseQA, 21\% lower in QQP,
and 9\% lower in SST.
Further, even though all negations are unimportant in WiC and WSC, 
we observe a drop in performance for the instances with negation compared to the instances without negation (WiC: 0.64 vs 0.67 and WSC: 0.59 vs 0.63). 
We conclude that transformers trained with existing NLU corpora face challenges with instances that contain negation.
These results raise two important questions for future research:
Is negation an inherently challenging phenomenon for RoBERTa?
How many instances with negation are required to solve a natural language understanding task?

\section{Conclusions}
\label{s:conclusions}
We have analyzed the role of negation in eight natural language understanding corpora covering six tasks.
Our analyses show
that~(a)~all but WSC contain almost no negations or around 
31\%--54\%
of the negations found in general-purpose texts,~(b)~the few negations in these corpora are usually unimportant,
and~(c)~RoBERTa obtains substantially worse results when negation is important.

Our analyses also provide some evidence that creating models to properly deal with negation may require both
new corpora and more powerful models.
The need for new corpora stems from the answers to Questions 1 and 2.
The justification for powerful models is more subtle.
We point out that the percentage of unimportant negations~(Section~\ref{s:experiments})
is only a weak indicator of the drop in results with important negations~(Table~\ref{t:results}).
For example, we observe a 24\% and 21\% drop in results with important negations from CommonsenseQA and QQP 
despite 45\% and 97\% of negations are unimportant. 

Negation reverses truth values thus solutions to any natural language understanding task
should be robust when negation is present and important.
To this end, our
future work includes two lines of research.
First, we plan to create benchmarks for the six tasks consisting of instances containing negation (50/50 split important/unimportant).
Second, we plan to conduct probing experiments to investigate whether (and where) pretrained transformers capture the meaning of negation.
Doing so may help us discover potential solutions to understand negation and make inferences.

\section*{Acknowledgements}
This material is based upon work supported by the National Science Foundation under 
Grant No.~1845757.
Any opinions, findings, and conclusions or recommendations expressed in this material
are those of the authors and do not necessarily reflect the views of the NSF.
The Titan Xp used for this research was donated by the NVIDIA Corporation.
Computational resources were also provided by the UNT office of High-Performance Computing. 
Further, we utilized computational resources from the Chameleon platform \cite{keahey2020lessons}. 
We also thank the reviewers for insightful comments.

\bibliography{refs}

\begin{thebibliography}{34}
\expandafter\ifx\csname natexlab\endcsname\relax\def\natexlab#1{#1}\fi

\bibitem[{Bar-Haim et~al.(2006)Bar-Haim, Dagan, Dolan, Ferro, Giampiccolo,
  Magnini, and Szpektor}]{bar2006second}
Roy Bar-Haim, Ido Dagan, Bill Dolan, Lisa Ferro, Danilo Giampiccolo, Bernardo
  Magnini, and Idan Szpektor. 2006.
\newblock The second pascal recognising textual entailment challenge.
\newblock In \emph{Proceedings of the second PASCAL challenges workshop on
  recognising textual entailment}, volume~6, pages 6--4. Venice.

\bibitem[{Bentivogli et~al.(2009)Bentivogli, Clark, Dagan, and
  Giampiccolo}]{bentivogli2009fifth}
Luisa Bentivogli, Peter Clark, Ido Dagan, and Danilo Giampiccolo. 2009.
\newblock The fifth pascal recognizing textual entailment challenge.

\bibitem[{Bowman et~al.(2015)Bowman, Angeli, Potts, and
  Manning}]{bowman-etal-2015-large}
Samuel~R. Bowman, Gabor Angeli, Christopher Potts, and Christopher~D. Manning.
  2015.
\newblock \href {https://doi.org/10.18653/v1/D15-1075} {A large annotated
  corpus for learning natural language inference}.
\newblock In \emph{Proceedings of the 2015 Conference on Empirical Methods in
  Natural Language Processing}, pages 632--642, Lisbon, Portugal. Association
  for Computational Linguistics.

\bibitem[{Cer et~al.(2017)Cer, Diab, Agirre, Lopez-Gazpio, and
  Specia}]{cer-etal-2017-semeval}
Daniel Cer, Mona Diab, Eneko Agirre, I{\~n}igo Lopez-Gazpio, and Lucia Specia.
  2017.
\newblock \href {https://doi.org/10.18653/v1/S17-2001} {{S}em{E}val-2017 task
  1: Semantic textual similarity multilingual and crosslingual focused
  evaluation}.
\newblock In \emph{Proceedings of the 11th International Workshop on Semantic
  Evaluation ({S}em{E}val-2017)}, pages 1--14, Vancouver, Canada. Association
  for Computational Linguistics.

\bibitem[{Dagan et~al.(2006)Dagan, Glickman, and
  Magnini}]{dagan:2005:PRT:2100045.2100054}
Ido Dagan, Oren Glickman, and Bernardo Magnini. 2006.
\newblock \href {https://doi.org/10.1007/11736790_9} {The pascal recognising
  textual entailment challenge}.
\newblock In \emph{Proceedings of the First International Conference on Machine
  Learning Challenges: Evaluating Predictive Uncertainty Visual Object
  Classification, and Recognizing Textual Entailment}, MLCW'05, pages 177--190,
  Berlin, Heidelberg. Springer-Verlag.

\bibitem[{Ettinger(2020)}]{ettinger-2020-bert}
Allyson Ettinger. 2020.
\newblock \href {https://doi.org/10.1162/tacl_a_00298} {What {BERT} is not:
  Lessons from a new suite of psycholinguistic diagnostics for language
  models}.
\newblock \emph{Transactions of the Association for Computational Linguistics},
  8:34--48.

\bibitem[{Geiger et~al.(2020)Geiger, Richardson, and
  Potts}]{geiger-etal-2020-neural}
Atticus Geiger, Kyle Richardson, and Christopher Potts. 2020.
\newblock \href {https://doi.org/10.18653/v1/2020.blackboxnlp-1.16} {Neural
  natural language inference models partially embed theories of lexical
  entailment and negation}.
\newblock In \emph{Proceedings of the Third BlackboxNLP Workshop on Analyzing
  and Interpreting Neural Networks for NLP}, pages 163--173, Online.
  Association for Computational Linguistics.

\bibitem[{Giampiccolo et~al.(2007)Giampiccolo, Magnini, Dagan, and
  Dolan}]{giampiccolo-etal-2007-third}
Danilo Giampiccolo, Bernardo Magnini, Ido Dagan, and Bill Dolan. 2007.
\newblock \href {https://www.aclweb.org/anthology/W07-1401} {The third {PASCAL}
  recognizing textual entailment challenge}.
\newblock In \emph{Proceedings of the {ACL}-{PASCAL} Workshop on Textual
  Entailment and Paraphrasing}, pages 1--9, Prague. Association for
  Computational Linguistics.

\bibitem[{Gururangan et~al.(2018)Gururangan, Swayamdipta, Levy, Schwartz,
  Bowman, and Smith}]{gururangan-etal-2018-annotation}
Suchin Gururangan, Swabha Swayamdipta, Omer Levy, Roy Schwartz, Samuel Bowman,
  and Noah~A. Smith. 2018.
\newblock \href {https://doi.org/10.18653/v1/N18-2017} {Annotation artifacts in
  natural language inference data}.
\newblock In \emph{Proceedings of the 2018 Conference of the North {A}merican
  Chapter of the Association for Computational Linguistics: Human Language
  Technologies, Volume 2 (Short Papers)}, pages 107--112, New Orleans,
  Louisiana. Association for Computational Linguistics.

\bibitem[{Hossain et~al.(2020)Hossain, Kovatchev, Dutta, Kao, Wei, and
  Blanco}]{hossain-etal-2020-analysis}
Md~Mosharaf Hossain, Venelin Kovatchev, Pranoy Dutta, Tiffany Kao, Elizabeth
  Wei, and Eduardo Blanco. 2020.
\newblock \href {https://doi.org/10.18653/v1/2020.emnlp-main.732} {An analysis
  of natural language inference benchmarks through the lens of negation}.
\newblock In \emph{Proceedings of the 2020 Conference on Empirical Methods in
  Natural Language Processing (EMNLP)}, pages 9106--9118, Online. Association
  for Computational Linguistics.

\bibitem[{Keahey et~al.(2020)Keahey, Anderson, Zhen, Riteau, Ruth, Stanzione,
  Cevik, Colleran, Gunawi, Hammock, Mambretti, Barnes, Halbach, Rocha, and
  Stubbs}]{keahey2020lessons}
Kate Keahey, Jason Anderson, Zhuo Zhen, Pierre Riteau, Paul Ruth, Dan
  Stanzione, Mert Cevik, Jacob Colleran, Haryadi~S. Gunawi, Cody Hammock, Joe
  Mambretti, Alexander Barnes, Fran\c{c}ois Halbach, Alex Rocha, and Joe
  Stubbs. 2020.
\newblock Lessons learned from the chameleon testbed.
\newblock In \emph{Proceedings of the 2020 USENIX Annual Technical Conference
  (USENIX ATC '20)}. USENIX Association.

\bibitem[{Khandelwal and Sawant(2020)}]{khandelwal-sawant-2020-negbert}
Aditya Khandelwal and Suraj Sawant. 2020.
\newblock \href {https://www.aclweb.org/anthology/2020.lrec-1.704}
  {{N}eg{BERT}: A transfer learning approach for negation detection and scope
  resolution}.
\newblock In \emph{Proceedings of the 12th Language Resources and Evaluation
  Conference}, pages 5739--5748, Marseille, France. European Language Resources
  Association.

\bibitem[{Kovatchev et~al.(2019)Kovatchev, Marti, Salamo, and
  Beltran}]{kovatchev-etal-2019-qualitative}
Venelin Kovatchev, M.~Antonia Marti, Maria Salamo, and Javier Beltran. 2019.
\newblock \href {https://doi.org/10.26615/978-954-452-056-4_067} {A qualitative
  evaluation framework for paraphrase identification}.
\newblock In \emph{Proceedings of the International Conference on Recent
  Advances in Natural Language Processing (RANLP 2019)}, pages 568--577, Varna,
  Bulgaria. INCOMA Ltd.

\bibitem[{Kwiatkowski et~al.(2019)Kwiatkowski, Palomaki, Redfield, Collins,
  Parikh, Alberti, Epstein, Polosukhin, Devlin, Lee, Toutanova, Jones, Kelcey,
  Chang, Dai, Uszkoreit, Le, and Petrov}]{kwiatkowski-etal-2019-natural}
Tom Kwiatkowski, Jennimaria Palomaki, Olivia Redfield, Michael Collins, Ankur
  Parikh, Chris Alberti, Danielle Epstein, Illia Polosukhin, Jacob Devlin,
  Kenton Lee, Kristina Toutanova, Llion Jones, Matthew Kelcey, Ming-Wei Chang,
  Andrew~M. Dai, Jakob Uszkoreit, Quoc Le, and Slav Petrov. 2019.
\newblock \href {https://doi.org/10.1162/tacl_a_00276} {Natural questions: A
  benchmark for question answering research}.
\newblock \emph{Transactions of the Association for Computational Linguistics},
  7:452--466.

\bibitem[{Levesque et~al.(2012)Levesque, Davis, and
  Morgenstern}]{10.5555/3031843.3031909}
Hector~J. Levesque, Ernest Davis, and Leora Morgenstern. 2012.
\newblock The winograd schema challenge.
\newblock In \emph{Proceedings of the Thirteenth International Conference on
  Principles of Knowledge Representation and Reasoning}, KR'12, page 552–561.
  AAAI Press.

\bibitem[{Liu et~al.(2019)Liu, Ott, Goyal, Du, Joshi, Chen, Levy, Lewis,
  Zettlemoyer, and Stoyanov}]{liu2019roberta}
Yinhan Liu, Myle Ott, Naman Goyal, Jingfei Du, Mandar Joshi, Danqi Chen, Omer
  Levy, Mike Lewis, Luke Zettlemoyer, and Veselin Stoyanov. 2019.
\newblock Roberta: A robustly optimized bert pretraining approach.
\newblock \emph{arXiv preprint arXiv:1907.11692}.

\bibitem[{Morante and Daelemans(2012)}]{morante-daelemans-2012-conandoyle}
Roser Morante and Walter Daelemans. 2012.
\newblock \href
  {http://www.lrec-conf.org/proceedings/lrec2012/pdf/221_Paper.pdf}
  {{C}onan{D}oyle-neg: Annotation of negation cues and their scope in conan
  doyle stories}.
\newblock In \emph{Proceedings of the Eighth International Conference on
  Language Resources and Evaluation ({LREC}'12)}, pages 1563--1568, Istanbul,
  Turkey. European Language Resources Association (ELRA).

\bibitem[{Naik et~al.(2018)Naik, Ravichander, Sadeh, Rose, and
  Neubig}]{naik-etal-2018-stress}
Aakanksha Naik, Abhilasha Ravichander, Norman Sadeh, Carolyn Rose, and Graham
  Neubig. 2018.
\newblock \href {https://www.aclweb.org/anthology/C18-1198} {Stress test
  evaluation for natural language inference}.
\newblock In \emph{Proceedings of the 27th International Conference on
  Computational Linguistics}, pages 2340--2353, Santa Fe, New Mexico, USA.
  Association for Computational Linguistics.

\bibitem[{Phang et~al.(2020)Phang, Yeres, Swanson, Liu, Tenney, Htut, Vania,
  Wang, and Bowman}]{phang2020jiant}
Jason Phang, Phil Yeres, Jesse Swanson, Haokun Liu, Ian~F. Tenney, Phu~Mon
  Htut, Clara Vania, Alex Wang, and Samuel~R. Bowman. 2020.
\newblock \texttt{jiant} 2.0: A software toolkit for research on
  general-purpose text understanding models.
\newblock \url{http://jiant.info/}.

\bibitem[{Pilehvar and
  Camacho-Collados(2019)}]{pilehvar-camacho-collados-2019-wic}
Mohammad~Taher Pilehvar and Jose Camacho-Collados. 2019.
\newblock \href {https://doi.org/10.18653/v1/N19-1128} {{W}i{C}: the
  word-in-context dataset for evaluating context-sensitive meaning
  representations}.
\newblock In \emph{Proceedings of the 2019 Conference of the North {A}merican
  Chapter of the Association for Computational Linguistics: Human Language
  Technologies, Volume 1 (Long and Short Papers)}, pages 1267--1273,
  Minneapolis, Minnesota. Association for Computational Linguistics.

\bibitem[{Rajpurkar et~al.(2016)Rajpurkar, Zhang, Lopyrev, and
  Liang}]{rajpurkar-etal-2016-squad}
Pranav Rajpurkar, Jian Zhang, Konstantin Lopyrev, and Percy Liang. 2016.
\newblock \href {https://doi.org/10.18653/v1/D16-1264} {{SQ}u{AD}: 100,000+
  questions for machine comprehension of text}.
\newblock In \emph{Proceedings of the 2016 Conference on Empirical Methods in
  Natural Language Processing}, pages 2383--2392, Austin, Texas. Association
  for Computational Linguistics.

\bibitem[{Ribeiro et~al.(2020)Ribeiro, Wu, Guestrin, and
  Singh}]{ribeiro-etal-2020-beyond}
Marco~Tulio Ribeiro, Tongshuang Wu, Carlos Guestrin, and Sameer Singh. 2020.
\newblock \href {https://doi.org/10.18653/v1/2020.acl-main.442} {Beyond
  accuracy: Behavioral testing of {NLP} models with {C}heck{L}ist}.
\newblock In \emph{Proceedings of the 58th Annual Meeting of the Association
  for Computational Linguistics}, pages 4902--4912, Online. Association for
  Computational Linguistics.

\bibitem[{Roemmele et~al.(2011)Roemmele, Bejan, and
  Gordon}]{roemmele2011choice}
Melissa Roemmele, Cosmin~Adrian Bejan, and Andrew~S Gordon. 2011.
\newblock Choice of plausible alternatives: An evaluation of commonsense causal
  reasoning.
\newblock In \emph{AAAI Spring Symposium: Logical Formalizations of Commonsense
  Reasoning}, pages 90--95.

\bibitem[{Socher et~al.(2013)Socher, Perelygin, Wu, Chuang, Manning, Ng, and
  Potts}]{socher-etal-2013-recursive}
Richard Socher, Alex Perelygin, Jean Wu, Jason Chuang, Christopher~D. Manning,
  Andrew Ng, and Christopher Potts. 2013.
\newblock \href {https://www.aclweb.org/anthology/D13-1170} {Recursive deep
  models for semantic compositionality over a sentiment treebank}.
\newblock In \emph{Proceedings of the 2013 Conference on Empirical Methods in
  Natural Language Processing}, pages 1631--1642, Seattle, Washington, USA.
  Association for Computational Linguistics.

\bibitem[{Stanovsky et~al.(2018)Stanovsky, Michael, Zettlemoyer, and
  Dagan}]{stanovsky-etal-2018-supervised}
Gabriel Stanovsky, Julian Michael, Luke Zettlemoyer, and Ido Dagan. 2018.
\newblock \href {https://doi.org/10.18653/v1/N18-1081} {Supervised open
  information extraction}.
\newblock In \emph{Proceedings of the 2018 Conference of the North {A}merican
  Chapter of the Association for Computational Linguistics: Human Language
  Technologies, Volume 1 (Long Papers)}, pages 885--895, New Orleans,
  Louisiana. Association for Computational Linguistics.

\bibitem[{Talmor et~al.(2019)Talmor, Herzig, Lourie, and
  Berant}]{talmor-etal-2019-commonsenseqa}
Alon Talmor, Jonathan Herzig, Nicholas Lourie, and Jonathan Berant. 2019.
\newblock \href {https://doi.org/10.18653/v1/N19-1421} {{C}ommonsense{QA}: A
  question answering challenge targeting commonsense knowledge}.
\newblock In \emph{Proceedings of the 2019 Conference of the North {A}merican
  Chapter of the Association for Computational Linguistics: Human Language
  Technologies, Volume 1 (Long and Short Papers)}, pages 4149--4158,
  Minneapolis, Minnesota. Association for Computational Linguistics.

\bibitem[{Wallace et~al.(2019)Wallace, Feng, Kandpal, Gardner, and
  Singh}]{wallace-etal-2019-universal}
Eric Wallace, Shi Feng, Nikhil Kandpal, Matt Gardner, and Sameer Singh. 2019.
\newblock \href {https://doi.org/10.18653/v1/D19-1221} {Universal adversarial
  triggers for attacking and analyzing {NLP}}.
\newblock In \emph{Proceedings of the 2019 Conference on Empirical Methods in
  Natural Language Processing and the 9th International Joint Conference on
  Natural Language Processing (EMNLP-IJCNLP)}, pages 2153--2162, Hong Kong,
  China. Association for Computational Linguistics.

\bibitem[{Wang et~al.(2019)Wang, Pruksachatkun, Nangia, Singh, Michael, Hill,
  Levy, and Bowman}]{NIPS2019_8589}
Alex Wang, Yada Pruksachatkun, Nikita Nangia, Amanpreet Singh, Julian Michael,
  Felix Hill, Omer Levy, and Samuel Bowman. 2019.
\newblock \href
  {http://papers.nips.cc/paper/8589-superglue-a-stickier-benchmark-for-general-purpose-language-understanding-systems.pdf}
  {Superglue: A stickier benchmark for general-purpose language understanding
  systems}.
\newblock In \emph{Advances in Neural Information Processing Systems 32}, pages
  3261--3275. Curran Associates, Inc.

\bibitem[{Wang et~al.(2018)Wang, Singh, Michael, Hill, Levy, and
  Bowman}]{wang-etal-2018-glue}
Alex Wang, Amanpreet Singh, Julian Michael, Felix Hill, Omer Levy, and Samuel
  Bowman. 2018.
\newblock \href {https://doi.org/10.18653/v1/W18-5446} {{GLUE}: A multi-task
  benchmark and analysis platform for natural language understanding}.
\newblock In \emph{Proceedings of the 2018 {EMNLP} Workshop {B}lackbox{NLP}:
  Analyzing and Interpreting Neural Networks for {NLP}}, pages 353--355,
  Brussels, Belgium. Association for Computational Linguistics.

\bibitem[{Warstadt et~al.(2019)Warstadt, Cao, Grosu, Peng, Blix, Nie, Alsop,
  Bordia, Liu, Parrish, Wang, Phang, Mohananey, Htut, Jeretic, and
  Bowman}]{warstadt-etal-2019-investigating}
Alex Warstadt, Yu~Cao, Ioana Grosu, Wei Peng, Hagen Blix, Yining Nie, Anna
  Alsop, Shikha Bordia, Haokun Liu, Alicia Parrish, Sheng-Fu Wang, Jason Phang,
  Anhad Mohananey, Phu~Mon Htut, Paloma Jeretic, and Samuel~R. Bowman. 2019.
\newblock \href {https://doi.org/10.18653/v1/D19-1286} {Investigating
  {BERT}{'}s knowledge of language: Five analysis methods with {NPI}s}.
\newblock In \emph{Proceedings of the 2019 Conference on Empirical Methods in
  Natural Language Processing and the 9th International Joint Conference on
  Natural Language Processing (EMNLP-IJCNLP)}, pages 2877--2887, Hong Kong,
  China. Association for Computational Linguistics.

\bibitem[{Williams et~al.(2018)Williams, Nangia, and Bowman}]{N18-1101}
Adina Williams, Nikita Nangia, and Samuel Bowman. 2018.
\newblock \href {http://aclweb.org/anthology/N18-1101} {A broad-coverage
  challenge corpus for sentence understanding through inference}.
\newblock In \emph{Proceedings of the 2018 Conference of the North American
  Chapter of the Association for Computational Linguistics: Human Language
  Technologies, Volume 1 (Long Papers)}, pages 1112--1122. Association for
  Computational Linguistics.

\bibitem[{Wu et~al.(2020)Wu, Wang, Yuan, Wu, and Li}]{wu-etal-2020-corefqa}
Wei Wu, Fei Wang, Arianna Yuan, Fei Wu, and Jiwei Li. 2020.
\newblock \href {https://doi.org/10.18653/v1/2020.acl-main.622} {{C}oref{QA}:
  Coreference resolution as query-based span prediction}.
\newblock In \emph{Proceedings of the 58th Annual Meeting of the Association
  for Computational Linguistics}, pages 6953--6963, Online. Association for
  Computational Linguistics.

\bibitem[{Yanaka et~al.(2019)Yanaka, Mineshima, Bekki, Inui, Sekine,
  Abzianidze, and Bos}]{yanaka-etal-2019-help}
Hitomi Yanaka, Koji Mineshima, Daisuke Bekki, Kentaro Inui, Satoshi Sekine,
  Lasha Abzianidze, and Johan Bos. 2019.
\newblock \href {https://doi.org/10.18653/v1/S19-1027} {{HELP}: A dataset for
  identifying shortcomings of neural models in monotonicity reasoning}.
\newblock In \emph{Proceedings of the Eighth Joint Conference on Lexical and
  Computational Semantics (*{SEM} 2019)}, pages 250--255, Minneapolis,
  Minnesota. Association for Computational Linguistics.

\bibitem[{Yang et~al.(2019)Yang, Wang, Liu, Liu, Lyu, Wu, She, and
  Li}]{yang-etal-2019-enhancing-pre}
An~Yang, Quan Wang, Jing Liu, Kai Liu, Yajuan Lyu, Hua Wu, Qiaoqiao She, and
  Sujian Li. 2019.
\newblock \href {https://doi.org/10.18653/v1/P19-1226} {Enhancing pre-trained
  language representations with rich knowledge for machine reading
  comprehension}.
\newblock In \emph{Proceedings of the 57th Annual Meeting of the Association
  for Computational Linguistics}, pages 2346--2357, Florence, Italy.
  Association for Computational Linguistics.

\end{thebibliography}
\bibliographystyle{acl_natbib}

\appendix

\section{Negation Cue Detection}
\label{s:negation_detection}

We develop a negation cue detector (Section \ref{s:experiments} in the paper) by utilizing the RoBERTa (base architecture; 12 layers) pre-trained model \cite{liu2019roberta}. 
We fine-tune the system on ConanDoyle-neg \cite{morante-daelemans-2012-conandoyle} corpus.
While fine-training, the negation cues are marked with BIO (B: Beginning of cue, I: Inside of cue, O: Outside of cue) 
tagging scheme. 
The contextualized representations from the last layer of RoBERTa are passed to a fully connected (FC) layer. 
Finally, a conditional random field (CRF) layer produces the output sequence for the labels.

Our model yields the following results on the test set:
93.26 Precision, 
94.32 Recall,
and
93.79 F1.
The neural model takes about two hours on average to train on a single GPU of NVIDIA Tesla K80.
A list of the tuned hyperparameters that the model requires to achieve the above results is provided in Table \ref{t:params_cue_detector}.
The code is available at \url{https://github.com/mosharafhossain/negation-and-nlu}.

\begin{table}[th!]
\centering
\begin{tabular}{lc}
\toprule
Hyperparameter           \\
\midrule
Max Epochs      &  50 \\
Batch Size      &  10 \\
Learning Rate (RoBERTa)   &  1e-5 \\
Learning Rate (FC, CRF)   &  1e-3 \\
Weight Decay (RoBERTa)    &  0.00001 \\
Weight Decay (FC)    &  0.001 \\
Grad Clipping   &  5.0 \\
Warmup Epochs   &  5 \\
Patience        &  15 \\
Dropout         &  0.5 \\
\bottomrule

\end{tabular}
\caption{Hyperparameters used to fine-tune the cue detector with ConanDoyle-neg \cite{morante-daelemans-2012-conandoyle} corpus. 
FC and CRF refers to fully connected and conditional random field layers, respectively.
}
\label{t:params_cue_detector}
\end{table}

\begin{table}[th!]
\centering
\begin{tabular}{lccc}
\toprule
                & Hp-1  &  Hp-2  &  Hp-3 \\
\midrule				
CmmnsnsQA	    &	10		   &   16         &   1e-5 \\
COPA            &   50         &   16         &   1e-5 \\
QQP             &   3          &   16         &   1e-5 \\
STS-B	        &	10		   &   16         &   1e-5 \\
QNLI	        &	3		   &   8          &   1e-5 \\
WiC	            &	10		   &   16         &   1e-5 \\
WSC	            &	200		   &   16         &   1e-6 \\
SST-2           &   3          &   16         &   1e-5 \\
\bottomrule

\end{tabular}
\caption{Hyperparameters used to fine-tune RoBERTa individually for each corpus. Hp-1, Hp-2, and Hp-3 refer to the number of epochs, batch size, and learning rate used in the training procedure. We use default settings for the other hyperparameters when we use the implementation by \newcite{phang2020jiant}.
}
\label{t:params_nlu_tasks}
\end{table}

\section{Hyperparameters to Fine-tune the System for Each of the NLU Tasks}
\label{s:generated_neg_pairs}

We use an implementation by \newcite{phang2020jiant} and fine-tune 
RoBERTa (base architecture; 12 layers) \cite{liu2019roberta} model separately for each of the eight corpora. 
We use the default settings of the hyperparameters, except for a few, when fine-tuning the model on each benchmark.
Table \ref{t:params_nlu_tasks} shows tuned hyperparameters for each benchmark.

\end{document}